\documentclass{article}

\usepackage[numbers, compress]{natbib}

    \usepackage[final]{neurips_2024}

\usepackage[utf8]{inputenc} 
\usepackage[T1]{fontenc}    
\usepackage{hyperref}       
\usepackage{url}            
\usepackage{booktabs}       
\usepackage{amsfonts}       
\usepackage{nicefrac}       
\usepackage{microtype}      
\usepackage{xcolor}         

\usepackage{amssymb,amsmath,amsthm}
\usepackage{tabularray}
\usepackage{graphicx}
\usepackage{subfig}
\usepackage{caption}
\usepackage{subcaption}
\usepackage{wrapfig}

\newtheorem{theorem}{Theorem}[section]
\newtheorem{lemma}[theorem]{Lemma}

\newtheorem{definition}[theorem]{Definition}

\usepackage[disable, textsize=tiny]{todonotes}
\newcommand{\hn}[1]{\todo{Hong: #1}}

\title{From  Models to Systems: 
 A Comprehensive Fairness Framework for Compositional Recommender Systems}

\author{%
  Brian Hsu \\
  LinkedIn Corporation \\
  Sunnyvale, CA \\
  \texttt{bhsu@linkedin.com} \\
  \And
  Cyrus DiCiccio \\
  USA \\
  \texttt{cjd48@cornell.edu} \\
  \AND
  Natesh S. Pillai  \\
  LinkedIn Corporation, Harvard University \\
  Sunnyvale, CA \\
  \texttt{napillai@linkedin.com} \\
  \And
  Hongseok Namkoong \\
  LinkedIn Corporation, Columbia Business School \\
  New York City, New York \\
  \texttt{namkoong@gsb.columbia.edu} \\
}

\begin{document}

\maketitle
\begin{abstract}
Fairness research in machine learning often centers on ensuring equitable performance of individual models. However, real-world recommendation systems are built on multiple models and even multiple stages, from candidate retrieval to scoring and serving, which raises challenges for responsible development and deployment. This system-level view, as highlighted by regulations like the EU AI Act, necessitates moving beyond auditing individual models as independent entities. We propose a holistic framework for modeling system-level fairness, focusing on the end-utility delivered to diverse user groups, and consider interactions between components such as retrieval and scoring models. We provide formal insights on the limitations of focusing solely on model-level fairness and highlight the need for alternative tools that account for heterogeneity in user preferences. To mitigate system-level disparities, we adapt closed-box optimization tools (e.g., BayesOpt) to jointly optimize utility and equity. We empirically demonstrate the effectiveness of our proposed framework on synthetic and real datasets, underscoring the need for a system-level framework.

\end{abstract}
\section{Introduction}
The prevailing focus in algorithmic fairness is on bias measurement and
mitigation for individual prediction models as the unit of
analysis~\cite{barocas-hardt-narayanan, mehrabi2022survey, caton2020fairness,
  wan2023survey}.  However, industrial applications of ML rarely train and
serve a single model in isolation: individual models are components of a
broader system.  The recent EU AI act defining its scope of a system as
"\textit{a machine-based system designed to operate with varying levels of
  autonomy and that may exhibit adaptiveness after deployment and that, for
  explicit or implicit objectives, infers, from the input it receives, how to
  generate outputs such as predictions, content, recommendations, or decisions
  that can influence physical or virtual environments}."  This system-level
perspective raises several questions for responsible development and
deployment of recommendation systems. Are fairness notions for individual
models sufficient to provide system-level equity?

\begin{figure}[t]
  \vspace{-1.5cm}
  \centering \includegraphics[width=\textwidth]{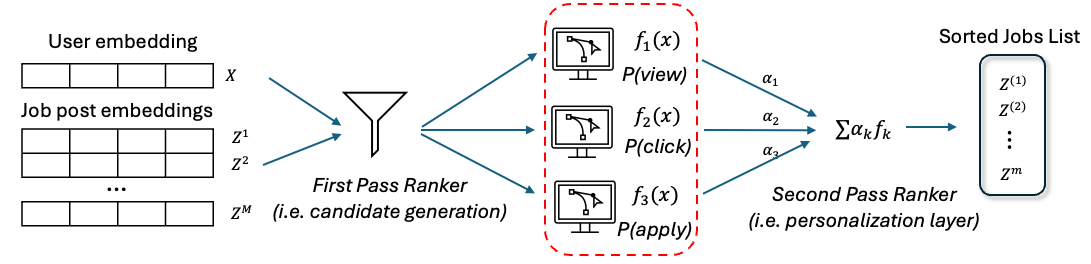}
      \caption{\label{fig:System} \textbf{AI Recommendation System Serving Pipeline.}
    Recommendations for feeds, ads, and social networking are generated from a
    multi-step process involving multiple ML models. An upstream process first
    fetches potentially relevant items via a candidate retrieval model, often
    called a “first pass ranker” (FPR), reducing the item set from millions to
    hundreds/thousands using scalable methods like approximate nearest
    neighbors~\cite{liu2004investigation}. Then, each model $f_{k}$ scores the
    items independently on based on probability of $\{$View, Click, Apply$\}$---this stage of ML models (red) are the overwhelming focus of fairness literature and audits. Finally, to surface the most relevant items, a "second pass ranker" (SPR) combines these individual models through a
    weighted sum $\sum_{k}\alpha_{k}f_{k}$.  While the SPR is simple, its intuitiveness has led to its
    ubiquity, with the world's largest platforms such as Meta~\cite{Instagram},
    LinkedIn~\cite{LinkedIn}, Microsoft~\cite{Microsoft},
    X/Twitter~\cite{Twitter}, Snapchat~\cite{Snapchat_spr},
    Pinterest~\cite{Pinterest}, and Spotify~\cite{Lamere_2021} stating or suggesting that they use a variant of this overarching system.} \vspace{-0.45cm}
\end{figure}

\begin{figure}[t]
    \centering
    \vspace{-1.5cm}
    {{\includegraphics[width=5.5cm, height=4.25cm]{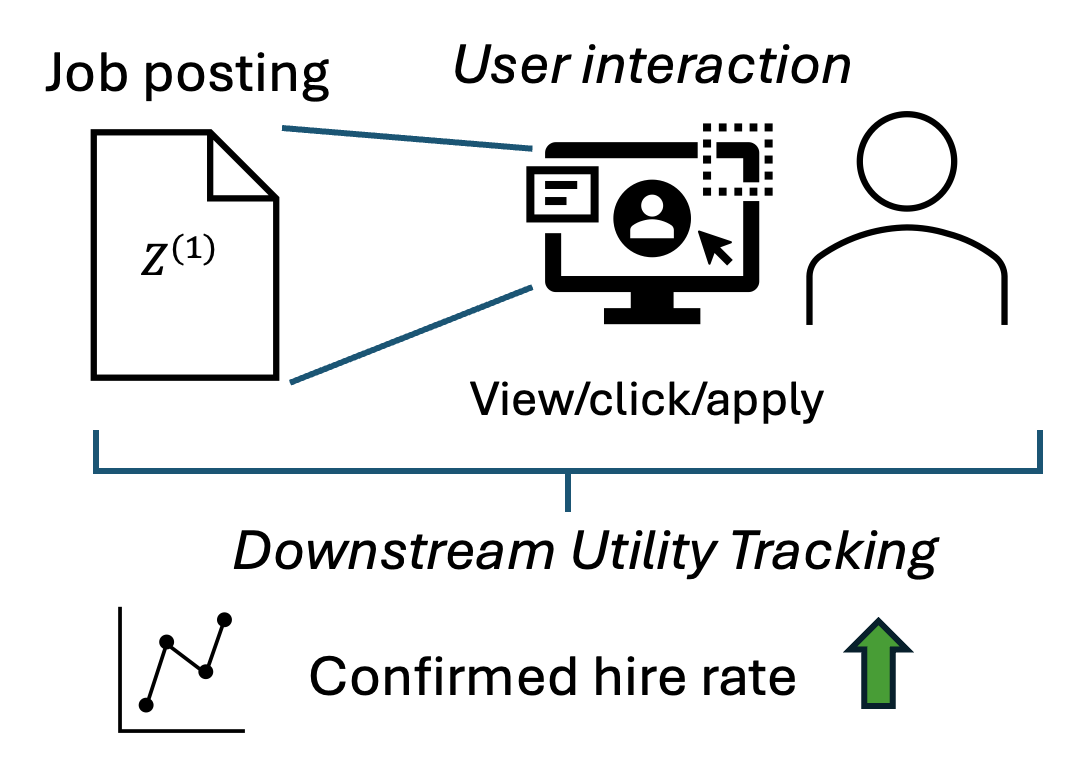} }}%
    \!
    {{\includegraphics[width=6cm, height=4.25cm]{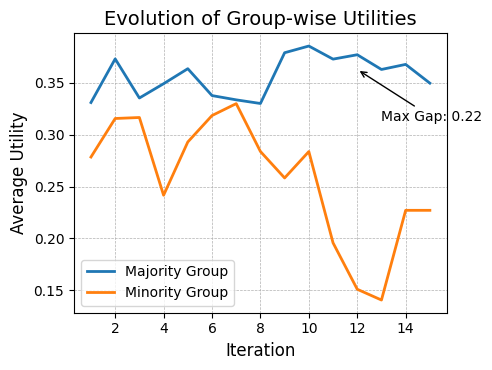} }}%
        \vspace{-5pt}
    \caption{(Left) Proxy and metric tracking (Right) Issues of vanilla BayesOpt } \label{fig: intro bayesopt}
    \vspace{-.5cm}
\end{figure}

We study system-level fairness in industrial recommendation systems, which are composed of multiple layers of ML models. In Figure \ref{fig:System}, we illustrate a schema utilized by many of the largest tech companies in the world. Using a job recommendation system as a running example, the ultimate goal is hiring, as measured by the ‘confirmed hire rate’ (see the left side of Figure \ref{fig: intro bayesopt}). However, this system-level objective is notably misaligned with the short-term predictions of individual models (e.g., clicks, views, applies). In turn, the fairness properties of those individual models may also be inadequate in addressing fairness at the system level. 
This discrepancy is exacerbated in practice since the aforementioned system design allows
going beyond individual model predictions to
integrate additional business logic such as priors on user preferences, known importance of different characteristics, prioritizing freshness of items.

We propose a system-level fairness framework for industrial recommendation engines that analyzes the entire pipeline, from retrieval to serving, rather than focusing on individual model fairness. This unified approach enables tracking bias propagation across components and its impact on utility disparities between user groups, particularly under preference heterogeneity ("distribution shifts"). We focus on the ultimate intent of the system, and using job recommendations as context, we develop a framework for measuring fairness through hiring outcome disparities.

\vspace{-0.25cm}
\paragraph{Multi-objective black-box optimization for system-level fairness}
In most settings, the performance of individual ML models can be optimized "off-line" based on previously collected user data using standard tools like cross-validation.
On the other hand, the FPR and SPR stages (Figure~\ref{fig:System}) require "online" experimentation (A/B testing) to collect user feedback on the quality of weights/parameters $\alpha$. In the latter, weights are typically tuned through manual trial \& error or black-box optimization methods such as Bayesian Optimization (BO)~\cite{frazier2018tutorial}. BO allows optimizing over noisy user feedback without derivative information, and is commonly used in online platforms (e.g., Meta~\cite{letham17} and LinkedIn~\cite{agarwal18}).
  
We go beyond identifying the source of  disparities in utility, and formulate the overall system design as a multi-objective optimization problem balancing social welfare and fairness. Since it is difficult to model system-level objectives using a particular functional form, we treat them as a black-box and model it using a flexible Gaussian process. We use a simple but effective variant of a Bayesian Optimization (BO) algorithm to  simultaneously maximize social welfare 
minus the disparity between the utility across groups. While the algorithms we leverage are well known, our formulation demonstrates how familiar tools from black-box optimization can be utilized in a practical and meaningful way to improve system-level fairness.

To further contextualize this in our job-recommendation setting, suppose that job preferences differ across demographic groups for jobs where one group prefers seeing attractive sounding jobs (click-worthy jobs), while another group prefers jobs of better background fit (apply-worthy jobs). When we have one model for clicks and one for applies and serve recommendations as a weighted linear combination of these model outputs, the "population weights" are unknown and are typically chosen "globally" in the sense that all users receive the same weights. Yet, the ideal weights from a user experience perspective largely depend on individual preference, which traditional fairness metrics fail to capture. While weights can be personalized to certain demographic groups, this introduces disparate treatment concerns~\cite{lipton2019does, seiner2006disentangling} and thus it is common practice to select a single global weight $\alpha$. As we demonstrate on the right side of Figure \ref{fig: intro bayesopt}, these global design choices have a tendency to tailor to the preferences of majority groups when using off-the-shelf tools. 

In Section~\ref{sec: bayesopt}, we formulate a black-box optimization problem over global weights that avoids starkly benefiting one group over others. While our specific fair BO methodology is straightforward, it underscores the critical role that adaptive experimentation methodologies like BO can play in achieving system-level fairness for industrial applications.

\vspace{-0.3cm}
\paragraph{Related work on viewer-side fairness in compositional systems} \label{sec: intro related works}
In this work, we focus on
\emph{viewer-side fairness}, which studies the disparity in utility a recommender system provides to users. In particular, this ignores the agency of the items being recommended: if items represent human agents and/or their products (e.g., creators and their content), a more holistic approach is required that goes beyond the scope of the current paper.

For fairness at an individual model-level, several authors focus on group-wise disparity of ranking performance metrics like AUC, NDCG, or F1, ensuring “minimum quality of service.” Examples include audits at Twitter (\cite{Lum_2022}), Microsoft (\cite{MicrosoftRAI}), and LinkedIn (\cite{Disentangling}), with comprehensive surveys by \citet{liRecSurvey}, \citet{wangRecSurvey}, \citet{AmifaMetrics}, and \citet{Ekstrand_2022}. We root our analysis on the spirit of viewer-side fairness concepts in this work and focus on the group-wise disparity in overall viewer \emph{utility}.

Fairness in compositional systems been studied from several perspectives; we summarize our key distinctions from existing works here and provide further technical details and references in Appendix \ref{appendix: related works}. Our work differs from existing research on compositional fairness in two key aspects: (1) we propose a novel multi-label fairness framework grounded in practical utility modeling, departing from prior approaches that propose transformations from the multi-label setting to a single-label setting \cite{Dwork2018FairnessUC, wang2021practical, MultiGroupCompositionalFairness}. (2) While fairness in ML pipelines has previously been studied \cite{bower2017fairpipelines, dwork2020individual, blum2022multi, khalili2021fair}, we take this one step further and jointly analyze the filtration/selection layer in conjunction with the serving layer. We use this perspective to emphasize the need for fairness interventions at this stage by showing how it can represent a bottleneck in achieving fairness, even if everything else is fully fair.

\paragraph{Limitations} Our main contribution is formulating and crystallizing the notion of fairness at the system-level, providing structural insights on how individual components can be optimized to achieve this goal. 
While formalizations provide practical value through concrete definitions and operational algorithms, our approach is inherently limited as system-level equity cannot be reduced to a single number. We highlight the need for multi-faceted approaches where i) stakeholders of the discrete components of the system collectively have incentives to prioritize equity at the system-level, and ii) a central organization coordinates system design with fairness as a central concern. Lastly, we make technical assumptions in Section \ref{sec: Framework} to focus on the single period interaction (rather than temporal) and focus on a specific variety of distribution shift. Handling these assumptions would introduce further granularity into the analysis that are tangential to our main message about system-level fairness and we consider these ideas as a topic for future study. 

\section{User Utility Under Compositional Recommender Systems} \label{sec: Framework}

We first argue that responsible AI in recommendation systems should prioritize downstream user utility over individual model performance metrics. Although previous works have advocated for utility-centric fairness (\cite{ZehlikeFairnessInRankingPartI}, \cite{ZehlikeFairnessInRankingPartII}, \cite{SinghRanking}), none have extended this to compositional systems where utility depends on multiple labels. Crucially, reducing disparities in single-model metrics (e.g., NDCG, AUC) may not improve system-level fairness, and proxy metrics used by individual models often misalign with the system's ultimate utility goal. Thus, we directly optimize for fairness in downstream utility rather than defining new composite metrics for translating fairness into the single-model domain.

\subsection{Framework for Compositional Utility}

We formalize the notion of system-level utility we explore in this work. In a recommendation system with $M$ items in its entire corpus (typically in the order of millions), let $Z^j$ be the feature representations of each item $j = 1, \ldots, M$. When a user initiates a session with covariates $X$, the item/candidate retrieval step selects $m < M$ relevant items; let
 $I(X,Z^{j}) \in \{0, 1\}$ be the indicator for whether $Z^{j}$ is selected. 
 In practice, item $Z^{j}$ may pass through rules-based filters or be chosen via a $m$-nearest neighbors algorithm.

For each item $j$ and user (query) $X$, we have $K$ different observable user outcomes $Y_k$ for $k = 1, \ldots, K$ ($K$ is usually $\le 10$, e.g., clicks, likes). These outcomes represent different aspects of item quality tracked by a domain expert (e.g., a product manager) who hypothesizes each $Y_{k}$ is a proxy of the downstream utility metric $U$ that they are tasked to improve (e.g., engagement, network growth). ML models $f_k: X, Z \rightarrow Y_{k}$ provide predictions of each outcome, which is combined with weights $\alpha_{k} > 0$ to recommend the best item $\hat{j}$ among the candidate set. 
\begin{definition}[Best Item for Serving] \label{def: item selection}
Given a universe of $M$ items, serving preferences $\alpha_{k} > 0$, model scores $f_{k}(X,Z^{j})$, and retrieval function $I(X,Z^{j})$, the recommended item $\hat{j}$ is given by
\begin{equation} \label{eq: argmax selection}
    \hat{j} := \mbox{argmax}_{1 \le j \le M}
    \left\{\sum_{k=1}^K \alpha_k f_k(X, Z^j) I(X,Z^{j})
    \right\}
\end{equation}
\end{definition}
\vspace{-5pt} \noindent 
Practitioners commonly use the weighted sum formulation~\eqref{eq: argmax selection} for interpretability and ease of adjustment. Even if the linear sum does not 
perfectly match user preferences, groups of users 
conceptually have an optimal set of weights 
within the linear model class. The universality of this model indicates that 
weighted sum is often viewed as a reasonable approximation of 
true user utility. Our framework generalizes to compositional functions (products, maxima), regression models, and settings involving negative utility (e.g., likelihood of abusive content) by simply adding more preference parameters. We focus on the top-1 item for ease of exposition; our analysis can be extended to top-k setting but in this case, modeling position bias across ranks $k$ is an important direction for future work. 

To formalize user utility, we give the system designer the utmost benefit of the doubt by assuming that their selection model is "well-specified". We shall see in the next section that even when we assume user utilities reflect the beliefs of domain experts, there can be large discrepancies between individual-model vs. system-level fairness.
\begin{definition}[User Utility] \label{def: user utility}
When item $\hat{j}$ is recommended, user $X$ utility is given by the conditional expectation of the weighted sum under true outcomes and true preferences  $\alpha^{*}(g)$ for group $ g\in \mathcal{G}$
\begin{equation} \label{eq: utility}
    U_{g}(X, I, f, \alpha, \alpha^{*})
    = \mathbb{E}
    \left[ 
    \sum_{k=1}^{K} \alpha_k^{\star}(g) \cdot Y^{\hat{j}}_{k}
    \mid X
    \right]
\end{equation}
\end{definition}
\vspace{-5pt}
\noindent Here, we implicitly assume $X$ is rich enough that $\{Y^j_k\} \mid X$ is invariant across demographic groups $g \in \mathcal{G}$ and instead model group-level heterogeneities through the true preference vector $\alpha^*(g)$.
Although users typically engage with the system multiple times, we treat them as i.i.d. in this work. We highlight this as a major limitation of our work as realistically, even a single bad experience can turn a user away from the platform. 

Using this formalization,  we can identify assumptions for when serving recommendations with $\alpha^{*}(g)$ is a maximizer of user utility. The conditions we require are that the scoring models $f_k$ are fair in the sense that they are  calibrated across the entire feature space. 
Calibration is an extensively studied topic in fairness literature~\cite{pmlr-v80-hebert-johnson18a, błasiok2023loss, pmlr-v97-liu19f, diciccio2022predictive}; though typically defined in case of candidate-side fairness, the same takeaways hold for the viewer-side case that we analyze.
\begin{lemma}
\label{lemma: optimal alpha}
Suppose individual models \( f_{k}(X,Z^{j}) \) are calibrated with respect to their intended label \( Y_{k} \) across the entire feature space:
$\mathbb{E}\left[Y_{k}^{j} \mid f_{1}(X, Z^{j}), \ldots, f_{K}(X, Z^{j})\right] = f_{k}(X,Z^{j})$ a.s. and that true and serving preferences are positive $\{\alpha^{*}, \alpha \} > 0$.
If item $j$ has a nonzero probability of being retrieved $\mathbb{P}\left(I(X,Z^{j}) =1 \right) > 0$ whenever $\mathbb{P}\left(f_k(X, Z^j) = \cdot \mid X, Z^j\right) > 0$ a.s., then setting $\alpha = c \cdot \alpha^*$ for $c > 0$ is a maximizer of utility.
\end{lemma}
\noindent The lemma shows the validity of our framework in that learning true preferences indeed optimizes utility. We also qualitatively consider the role and impact of \textit{unobservable} outcomes (e.g. sentimentality) in Appendix \ref{appendix: unobservables}.

\section{Structural Insights}

We now provide structural analyses of the utility gap between groups to answer the following question: \textit{when is individual model fairness (in)sufficient for system-level fairness?} 
We articulate the different sources of utility-based inequity, and show the causes for utility gaps can be different compared to the standard causes of individual-model fairness. In particular, users who have the same features $X$ look identical to the platform, but they may exhibit heterogeneous preferences (and therefore utilities) across groups. For example, different demographic groups may not respond identically to the same job posting despite similar backgrounds. 

We consider two groups $G = 0$ and $G = 1$ to illustrate, and use $E_{g}[\cdot]$ to denote the expectation with respect to $\mathbb{P}(X = \cdot |G=g)$. Focusing on the user's short-term experience after a single interaction with the recommendation system, we analyze the utility gap
    $$\mathbb{E}_{1} \left[ U_{1}(X, I, f, \alpha, \alpha^{*}) \right] - \mathbb{E}_{0}\left[ U_{0}(X, I, f, \alpha, \alpha^{*}) \right].$$
We assume the product \textit{does not} personalize models specifically to demographic groups at any stage due to disparate treatment concerns---treating users differently based on their demographics. Without loss of generality, we denote group $0$ to be the disadvantaged group. 

Our goal is to perform an apples-to-apples comparison between demographic groups by "controlling" for the effects due to the user features $X$. 
That is, we wish to compare
the gap utility $U_{1}(X, I, f, \alpha, \alpha^{*})  -  U_{0}(X, I, f, \alpha, \alpha^{*})$
for users that only differ in their group memberships. However, such a comparison is only possible over users co-observed in both groups.
Thus, we define a notion of a "shared space" between $\mathbb{P}(X | G = 1)$ and $\mathbb{P}(X | G = 0)$
\begin{equation}
    \label{eqn:shared}
    S_X(x) \propto (p(x | G = 1) + p(x | G = 0))^{-1} p(x | G = 1) p(x | G = 0),
\end{equation}
so that $S_X$ is small whenever either $p(x | G = 1)$ or $p(x | G = 0)$ is small and large when both quantities are large.
Intuitively, taking expectations over $S_{X}$ means we are paying attention to the feature space where both groups are present (e.g. industries $X_1$ where both demographics $G = 0, 1$ are represented). This mirrors \citet{cai2023diagnosing}'s distribution shift decomposition approach.

We expand the difference in expected utility as follows and provide some high level intuition:
\begin{align}
    \mathbb{E}_{1} & \left[ U_{1}(X, I, f, \alpha, \alpha^{*}) \right] - \mathbb{E}_{0} \left[ U_{0}(X, I, f, \alpha, \alpha^{*}) \right] \nonumber \\
    &= \mathbb{E}_{1}\left[ U_{1}(X, I, f, \alpha, \alpha^{*}) \right]  - \mathbb{E}_{S_{X}}[U_{1}(X, I, f, \alpha, \alpha^{*})] \label{eq:x shift 1} \\
    & \qquad + \mathbb{E}_{S_{X}} \left[ U_{1}(X, I, f, \alpha, \alpha^{*}) - U_{0}(X, I, f, \alpha, \alpha^{*}) \right] \label{eq:yx shift} \\
    & \qquad  + \mathbb{E}_{S_{X}}[U_{0}(X, I, f, \alpha, \alpha^{*})] - \mathbb{E}_{0}\left[ U_{0}(X, I, f, \alpha, \alpha^{*}) \right] \label{eq:x shift 2}
\end{align}
 
 Term~\eqref{eq:x shift 1} relates to the utility change from the feature distribution of group $1$ ($P(X|G=1)$) to the shared distribution~\eqref{eqn:shared}. It is large when the utility gap can be attributed to user features $X$ often seen in group 1 but not in group 0. 
 For instance, this may imply that the AI system provides better recommendations in male-dominated industries such as construction. 
 
Term~\eqref{eq:yx shift} compares the utility gap between groups over $S_{X}$. This term is large when the AI system favors the preferences $\alpha^{*}$ of the majority group 1.  
Finally, Term~\eqref{eq:x shift 2} is large
if the AI system works better for users who are common in both groups compared to those only in group 0.

\subsection{Impact of Preference Misspecification} \label{subsection: preference misspecification}

We now take a closer look at Term~\eqref{eq:yx shift}. Further decomposition of these terms in Theorems~\ref{lemma: alpha shift},~\ref{lemma: CR dist shift} to come unveil the impact and limitations of individual model fairness. We show that misspecification of user preferences in the serving model and disparities in the quality of the embedding model can also be significant drivers of utility gaps.

We analyze the "apples-to-apples" comparison in Term \eqref{eq:yx shift}, which provides the most intuitive notion of "unfairness." This implies that the quality of recommendations is unequal even when two  users from different groups share the same covariates (e.g. in interest and past behavior). We show this gap can occur at the system-level \textit{even if the individual models are fair}.
Below, the residuals $|Y_{k}^{\hat{j}}-f_{k}(X,Z^{\hat{j}})|$ measure model performance of $f_k$, where as $|\alpha_{k}-\alpha_{k}^{*}(0)|$ and $|\alpha_{k}-\alpha_{k}^{*}(1)|$ denote the estimation error in preferences of the two groups.
\begin{theorem}[Utility Gap Bound From Preference Misspecification] \label{lemma: alpha shift}
    If recommendations are served using one set of $\alpha$'s for both groups, the expected utility gap (Term \ref{eq:yx shift}) is upper bounded by
    \begin{align*}
    \sum_{k}
    \mathbb{E}_{S_{X}} \biggl[ |\alpha^{*}_{k}(1)-\alpha_{k}^{*}(0)| \cdot |Y_{k}^{\hat{j}}-f_{k}(X,Z^{\hat{j}})| 
     + |\alpha_{k}-\alpha_{k}^{*}(0)| \cdot f_{k}(X,Z^{\hat{j}}) + |\alpha_{k}-\alpha_{k}^{*}(1)| \cdot f_{k}(X,Z^{\hat{j}})  \biggr].
\end{align*}
\end{theorem}
\vspace{-5pt}
Typical fairness interventions aim to reduce the first term by ensuring each individual model performs well~\cite{pmlr-v80-hebert-johnson18a, błasiok2023loss, gopalan2021omnipredictors}. Our bound affirms individual model fairness is a crucial requirement, but also that disparities in model performance are amplified by differences in preferences across groups $\alpha_{k}^{*}(1)-\alpha_{k}^{*}(0)$, which are terms \textit{not} reducible by the modeler. 
On the other hand, the second and third terms represent the utility disparity caused by misspecification of preferences. 
The utility disparity due to heterogeneous preferences has been been largely left out in the fairness literature; even works on compositional fairness treat all models/labels as equally important. We propose a concrete algorithmic approach in Section \ref{sec: bayesopt} to address this problem.  

\subsection{Downstream Impact of Upstream Candidate Selection Models} \label{subsection: Retrival distribution shift}

We now focus on Term \eqref{eq:x shift 1}, and move \textit{upstream} of the scoring models to assess the impact of candidate retrieval model $I(X,Z)$ as a driver of utility gap.
In practice, the candidate retrieval model is separate from the ML models and SPR serving layer and may even be managed by a separate engineering team. Retrieval models are often designed to optimize offline retrieval metrics such as recall over a single outcome (e.g., Click).
Since engineering considerations such as latency, memory, and performance drive retrieval model design~\cite{Pinterest_fpr, Uber_fpr, Snapchat_fpr}, fairness considerations are generally underappreciated.

While understanding user preferences over multiple labels is key to maximizing utility, candidate retrieval evaluations do not typically consider multiplicity of labels: recall over a single label may not align with the important label from the user's perspective. We address this by proposing a retrieval quality metric that we find more suitable for the compositional model system. 
\begin{definition}[Candidate Retrieval Model Quality]
    A $\gamma$-good item $j$ satisfies $\mathbb{E}\left[\sum_{k=1}^{K} Y_{k}^{j}\right] \geq \gamma$. The quality of a candidate retrieval model $I(X, Z)$  selecting $m$ items is the expected highest $\gamma$-good item it can retrieve from the candidate pool.
    Formally, recalling $I(X, Z^j)$ is the indicator for whether item $Z^j$ is retrieved for user $X$, the quality metric for the candidate retrieval model is
    \[
    Q_{m}(I(X,Z), Y) = \mathbb{E}\left[\max_{j \in \{1, \ldots, m\}} \left( I(X, Z_j) \cdot \sum_{k=1}^{K} Y_{k}^{j} \right)\right]
    \]
\end{definition}
Notably, this definition differs from the standard definition of recall (true positives over total positives). This distinction is crucial for two reasons. First, this definition now spans multiple labels that the business ultimately knows are relevant for user preferences. Second, we focus on the maximum because the downstream model and SPR layer are designed for surfacing the best item from the retrieved candidates. 

With this definition in hand, we relate disparity in retrieval quality to that of utilities.
We show that even if fully optimize utility with respect to everything \textit{downstream} of the candidate retrieval model $I(X,Z)$, namely the scoring models $f_k$ and serving coefficients $\alpha_k$---the utility gap is still bottlenecked by the quality of the candidate retrievals.
\begin{theorem}[Utility Gap Bound From Candidate Retrieval Performance Degradation] \label{lemma: CR dist shift}
Assume away biases from other sources so that $\alpha=\alpha^{*}$ and individual models are calibrated as in Lemma \ref{lemma: optimal alpha}. 
    For all users, let there be $m^{+}$ $\gamma$-good items in our item corpus of size $M$, and consider a $\epsilon$-gap in retrieval quality
    $\mathbb{E}_{1} \left[Q_{m}(I(X,Z), Y) \right] - 
    \mathbb{E}_{S_{X}}\left[Q_{m}(I(X,Z), Y) \right]
    > \epsilon$. 
     Then, Term~\eqref{eq:x shift 1} is at least
    \[
     \sup_{f, \alpha} \mathbb{E}_{1} \left[ U_{1}(X, I, f, \alpha, \alpha^{*}) \right] - \sup_{f, \alpha} \mathbb{E}_{S_{X}} \left[ U_{1}(X, I, f, \alpha, \alpha^{*}) \right] \geq \epsilon \cdot \min_{k} \alpha^{*}_{k}(1).
    \]\hn{I don't understand what the previous sup is over. Do you actually need it? {BH} lol honestly, it's not a great reason - the assumptions give expectation bounds that look like 0 <= E <= U, so taking the difference of two upper bounds does not give a nice lower bound unless we difference the sup of the upper bounds and then lower bound that.}
\end{theorem}

\section{System-Level Fairness Via Bayesian Optimization} \label{sec: bayesopt}

We now shift our focus from identification of fairness gaps to its mitigation. Based on our discussion in Section \ref{subsection: preference misspecification}, a significant contributor of the utility gap is over-representation of the majority group in selecting the preference weights $\alpha$. 
We formulate the selection of $\alpha$ as a derivative-free closed-box optimization and propose
a mitigation strategy based on inequality-aware Bayesian optimization (BO). Beyond our specific methodology, our main insight is that adaptive experimentation methodologies like BO have great potential in delivering system-level fairness.

\subsection{Selecting an Inequality Metric}
Prior works at the intersection of BO and fairness have generally centered on finding hyperparameters that yield a fair model based on a standard fairness definition such as demographic parity or equalized odds (see \cite{perroneFairBayesOpt}, \cite{Weerts_2024}, \cite{candelieri2022fair}, \cite{SikdarGetfair} for examples). However, we are specifically proposing to forgo these standard definitions in favor of utility-based fairness. While it becomes tempting to directly take the exact term we analyzed (average utility gap across the groups) as a measure of inequality, there are various issues such as not being scale-invariant, the need to designate a specific disadvantaged group, and inability to compare across more than two groups, that make it unsuitable for our application. 

Instead, we leverage the recently proposed metric Deviation from Equal Representation (DER).
\citet{friedberg2022representation} introduce the metric as a way of quantifying the disparate impact of experiments.  For $k$ groups with non-negative downstream outcomes $\mu_{1}, \cdots, \mu_{k}$ (representing average number of sessions/confirmed hires per group), the DER is defined 
\footnote{Although DER is originally defined as 1 minus the quantity shown, we renormalize it to have the interpretation that "higher is more equitable/better" to better align with the canonical view of Bayesian optimization.} as the following metric, which is scale invariant and naturally extends to multiple groups.
\begin{equation}
\label{eq:der}
D(\mu_{1}, \cdots, \mu_{k}) = 1- \frac{k}{k-1} \sum_{k}( \frac{\mu_{k}}{\sum_{k}\mu_{k}} - \frac{1}{k})^{2} 
\end{equation}

 When all means are equal, $D(\mu_{1},\mu_{2})=0$ and otherwise the statistic grows larger when the values grow more disparate. \citet{friedberg2022representation} uses this for gauging if an experiment has unintended consequences (e.g., disproportionately benefits one group) that are not identifiable by looking at a global metric change. Similarly, our system-level fairness perspective aims to ensure the overall utility gains are not due to a disproportionate gain in the majority group and a drop in minority group. We use DER to integrate fairness considerations into the Bayesian optimization process.

\subsection{Incorporating Deviation from Equal Representation Into Bayesian Optimization for System-Level Fairness}

A common strategy to considering fairness in BO is to use a constrained version of the expected improvement (EI) methodology by ~\citet{gardner2014bayesian}. Here, the performance and fairness criterion are parameterized as separate Gaussian Processes (GP) over the decision variable $\alpha$ (preference weights). We briefly recap these strategies and explain why we opt for a different, but related procedure. Originally, the EI approach~\cite{frazier2018tutorial} selects the next point optimizing the following quantity at step $n$ 
\[
EI_{n}(\alpha) = \mathbb{E}_{n}\left[ [f(\alpha)-f_{n}^{*}]^{+} \right].
\]
The "fair variant" of the algorithm modifies this objective by introducing an indicator variable $I(\alpha)$ parameterized by a GP $c(\alpha)$ to represent if a point $\alpha$ satisfies the constraints or not. Then, by assuming independence between the constraint and objective, \citet{gardner2014bayesian} presents the constrained objective $EIC_{n}$ where $\gamma$ is a hyperparameter that denotes the slack on the constraint $c(\alpha) -\gamma \le 0$. 
\begin{align*}
    EIC_{n}(\alpha) & = \mathbb{E}_{n}\left[ [f(\alpha)-f_{n}^{*}]^{+} \cdot I(\alpha) \right] 
     = \mathbb{E}_{n}\left[ [f(\alpha)-f_{n}^{*}]^{+}  \right] \cdot \mathbb{E}_{n}\left[ I(\alpha)  \right] 
     = EI_{n}(\alpha)\cdot \mathbb{P}(c(\alpha) \leq \gamma)
\end{align*}
This formulation is interpretable and maintains all the computational advantages of the standard EI method. Prior works demonstrate the efficacy of this method for finding hyperparameters that optimize for performance while satisfying fairness constraints \cite{perroneFairBayesOpt}. However, the problem with directly applying this method by parameterizing DER~\eqref{eq:der} as a GP $c(\alpha)$ is twofold. First, DER clearly is not independent of the original objective of maximizing the overall downstream metric. This voids the mathematical support behind the constrained EI. Second, picking a threshold $\gamma$ for DER is nontrivial. Realistically, the business is more interested in finding the best of both worlds by improving DER and the overall average utility metric simultaneously.

Hence, rather than rely on the constrained EI formulation, we propose treating the problem as a multi-objective-optimization problem and using Expected Hyper-Volume Improvement (EHVI). As its name suggests, EHVI is the multi-objective analog of EI, where instead of optimizing for improvement in a scalar value, we now look to improve upon the hypervolume of the Pareto frontier. We include the definition from \citet{daulton2020ehvi} below for completeness; see \citet{daulton2020ehvi, yangehvi} for more background on the method.
\begin{definition} (EHVI)
\hn{I tried to translate notation to ours---please double check.{BH} better to use a different symbol for the reference point as it's a fixed value (usually the origin (0.0, 0.0)}
Given a reference point $r \in \mathbb{R}^{K}$, the hypervolume indicator of a Pareto set $\mathcal{P}$ is the $K$-dimensional Lebesgue measure $\lambda_K$ of the space dominated by $\mathcal{P}$ and bounded from below by $\alpha$: $HV(\mathcal{P}, r)=\lambda_{K} \left( \bigcup_{i=1}^{|\mathcal{P}|} \left[r, y_{i} \right] \right)$ where $\left[r, y_{i} \right]$ denotes the hyperrectangle bounded by vertices $r$ and $y_{i}$.    
\end{definition}

By optimizing for the global utility and DER simultaneously, we aim to find points that present the best possible tradeoffs. At each iteration, EHVI returns a set of preference weights that have the highest expectation of hypervolume improvement and we prioritize those with the highest expected DER. This allows us to find points that encourage fairness while also boosting global utility.

\section{Experiments}
\subsection{Datasets}

\begin{wrapfigure}{R}{0.45\textwidth}
  \centering
  \vspace{-1.25cm} 
  \includegraphics[width=.9\linewidth]{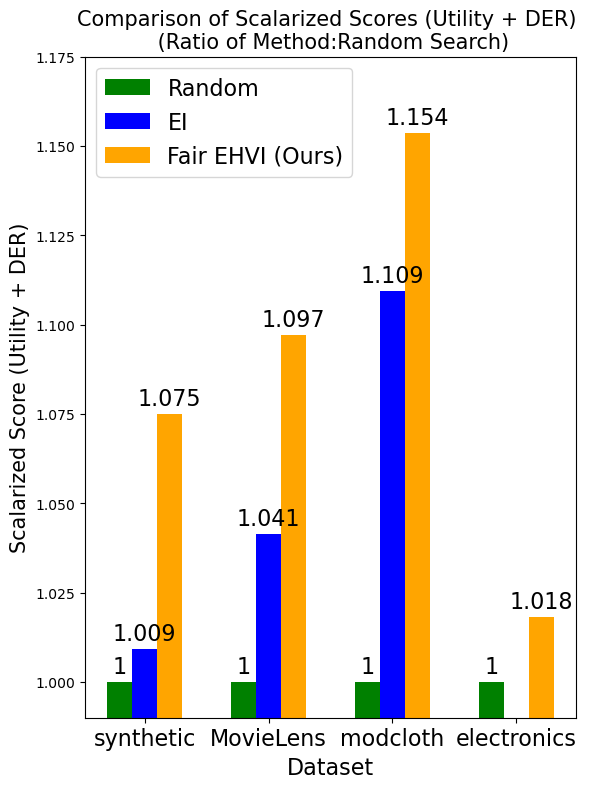} 
  \vspace{-0.2cm} 
  \caption{Comparison of methods on scalarized outcome}%
  \label{fig:barplot_comparison}%
  \vspace{-0.5cm} 
\end{wrapfigure}

We now demonstrate our proposed multi-objective-optimization formulation of utility and DER as a simple but effective method for achieving system-level fairness. Due to the lack of public benchmarks with readily available multi-action outcomes to credibly represent an industrial system, we restrict our analysis to four datasets \texttt{MovieLens}(\cite{harper2015movielens}), \texttt{ModCloth}(\cite{wan2019addressingmarketingbiasproduct}), \texttt{Electronics}(\cite{wan2019addressingmarketingbiasproduct}), and one synthetic example. These datasets represent varying levels of difficulty in terms of model performance, utility optimization, and DER optimization. Details are provided in Appendix \ref{appendix: data}. Importantly, we \textit{start} with models that are groupwise fair, which is typically the end-goal of fairness studies. We then simulate group-wise preference disparities and use BOTorch \cite{balandat2020botorch} for iterative Bayesian optimization to mimic online recommendation systems. Each iteration samples users, retrieves $m$ items per user, and computes utility based on true labels and preferences as defined in Definition \ref{def: user utility}.

\subsubsection{Results and Comparison}
For each dataset and method, we aggregate the utility and DER over the 20 iterations for each of the 20 trials. We trace the Pareto frontier for each method across the trials, and then plot the average utility and DER to represent the average-case performance with 1 stdev error bars. For statistical significance, we report the p-value of the Wilcoxon signed rank test comparing our Fair EHVI against the baseline (random search) and pure utility optimization (EI). We consider random search the baseline, as even simple BO processes like EI are non-trivial implement in industry and therefore may not be available. All experiments were run locally on a MacBook Pro with an M3 processor.



\begin{figure}[t]
\vspace{-1cm} 
    \centering
    \begin{minipage}[t]{0.49\textwidth}
        \centering
        \includegraphics[width=\textwidth, height=5.5cm]{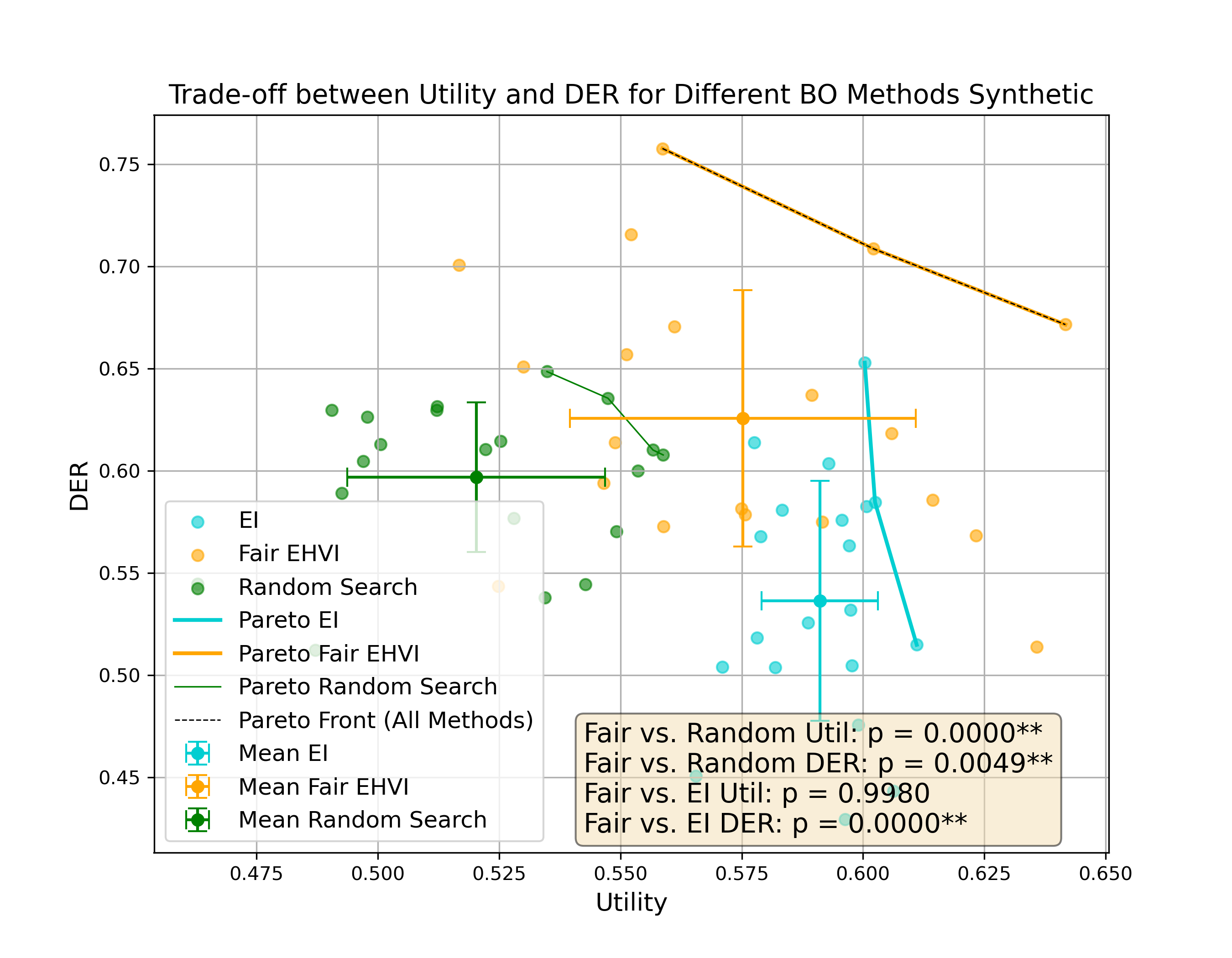}
    \end{minipage}
    \hfill
    \begin{minipage}[t]{0.49\textwidth}
        \centering
        \includegraphics[width=\textwidth, height=5.5cm]{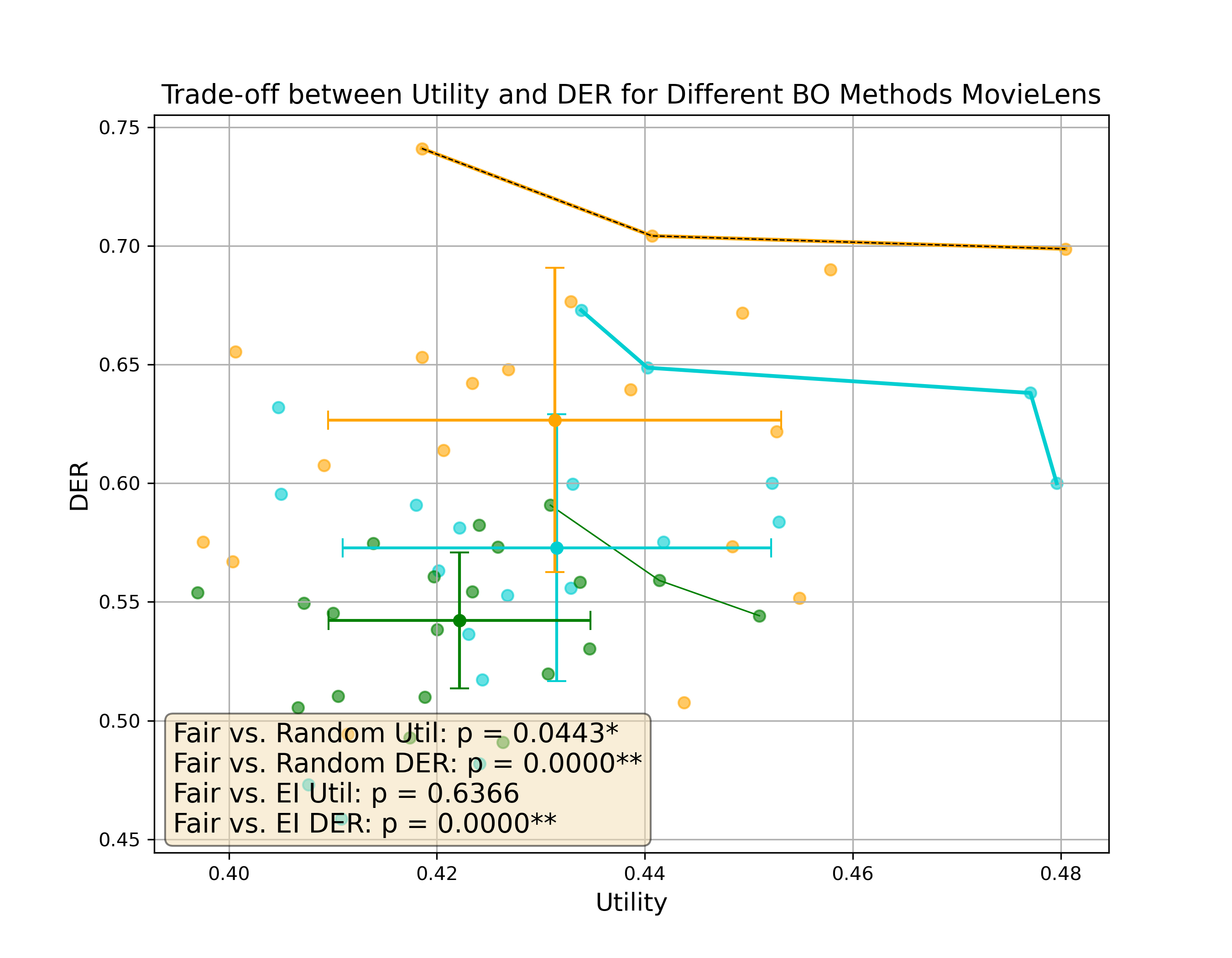}
    \end{minipage}
    
    \vspace{-0.25cm} 

    \begin{minipage}[t]{0.49\textwidth}
        \centering
        \includegraphics[width=\textwidth, height=5.5cm]{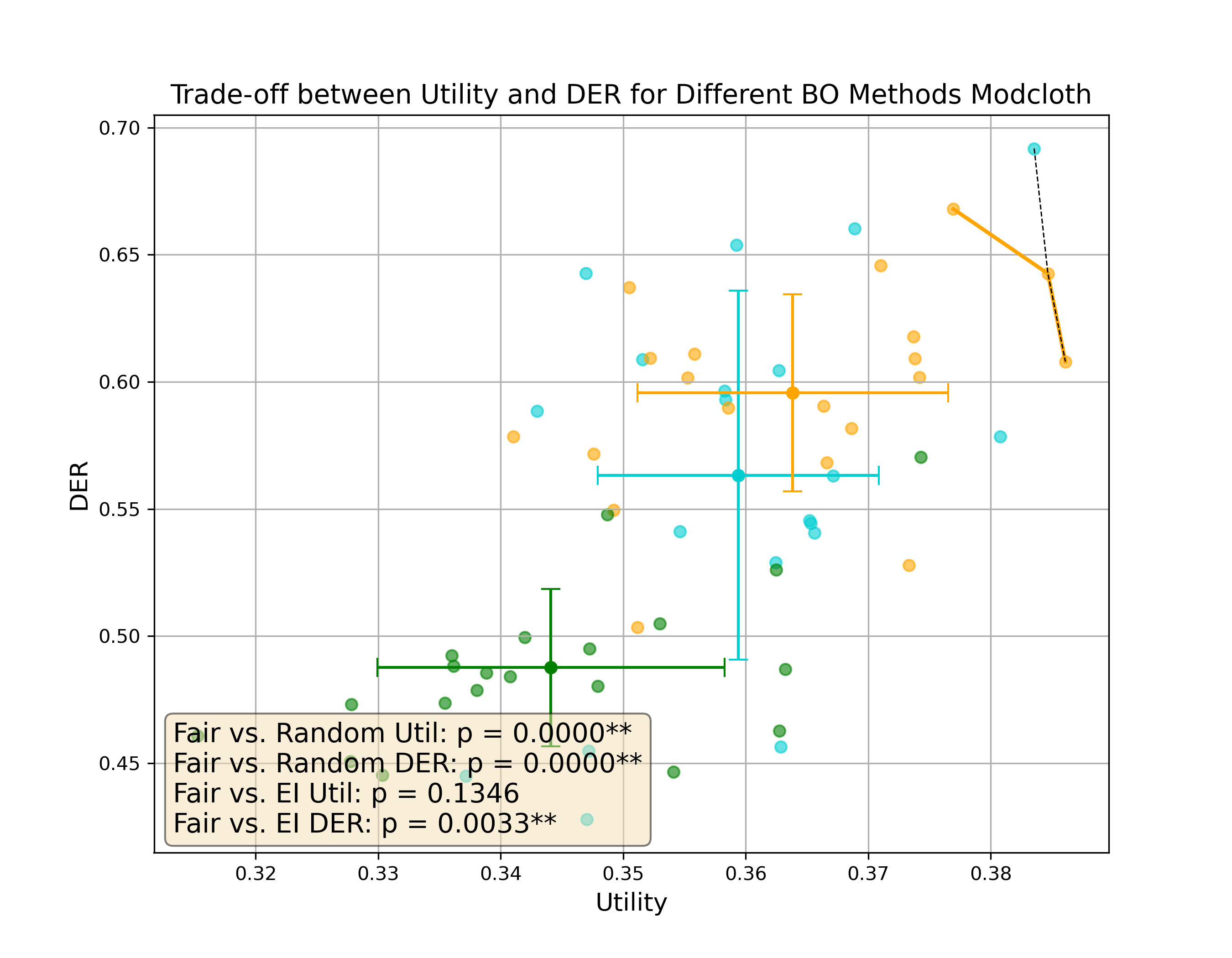}
    \end{minipage}
    \hfill
    \begin{minipage}[t]{0.49\textwidth}
        \centering
        \includegraphics[width=\textwidth, height=5.5cm]{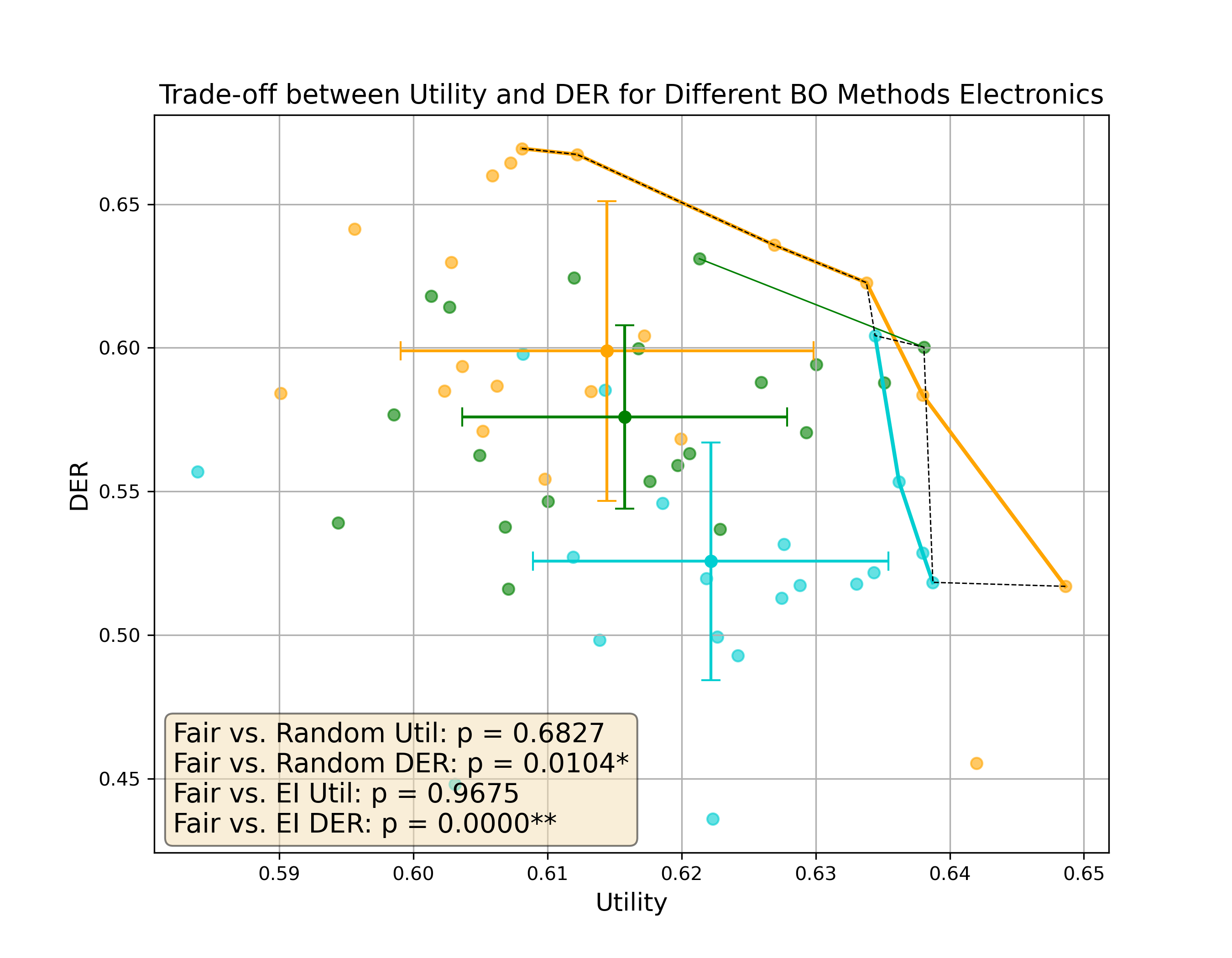}
    \end{minipage}
    
    \vspace{-0.4cm} 
    \caption{Pareto frontiers for the four tested datasets}
    \label{fig:pareto_frontiers}
\end{figure}

In Figure \ref{fig:barplot_comparison}, we see that Fair EHVI overall beats both random search and EI in identify the best tradeoff between utility and DER. For a more detailed view of performance on each dataset, we turn to Figure \ref{fig:pareto_frontiers}. The plots demonstrate that Fair EHVI consistently yields the most points on the global pareto frontier of utility and DER. Fair EHVI also yields better DER than random search and EI by a statistically significant margin in all cases, even in the difficult Electronics dataset where random search is difficult to beat. Our methodology does suffer in that dataset where the average utility is slightly lower than random search, but this can potentially be overcome with more tuning of the search parameters (as our algorithm samples EHVI candidates in parallel and optimizes for DER). Overall, we reiterate that the significance of adaptive experimentation tools like BO in AI systems necessitates further research on using them to trade between fairness and business objectives, or even between fairness objectives (\cite{Hsu2022}, \cite{bell2023possibility}).

\section{Conclusion and Future Work}

In this paper, we have proposed a mechanism for shifting from model-centric to system-level fairness in compositional recommender systems. We align fairness measurement with user utility optimization, spotlight the retrieval and serving layers as critical points of intervention, demonstrate the benefits and limitations of intervention at these layers, and propose a Bayesian optimization solution for bias mitigation at serving time. Our empirical results show improved utility distribution across heterogeneous user groups. As regulations like the EU AI Act emerge, such holistic system-level analysis becomes crucial for responsible AI practices. 

For future work, one critical direction is understanding how the timescales of utilities affect fairness. For instance, while we have assumed that either all users are unique or that user sessions are i.i.d, this is generally not true. High utility in one session begets further usage and low utility begets strong drop-off. While these patterns can only be empirically estimated, folding them into our fairness framework adds the timescale dimension of how long unfairness is tolerable by users, further motivating the urgency of the problem. Additionally, as we have framed system-level fairness for the viewerside in this paper, we encourage researchers to understand the analagous problem but for systems that rank candidates and the interaction between the two types of fairness. 

\subsection{Acknowledgments}

We sincerely thank Sam Gong and Will Cai for their insightful feedback and in-depth conversations about system-level fairness. We would also like to thank Sakshi Jain and Heloise Logan for their support and Kinjal Basu for his improvements on the DER metric. Finally, we thank the anonymous reviewers for their helpful comments on notation improvements and for providing references to expand the discussion of related works.

\newpage
\appendix
\section{Appendix / Related Works} \label{appendix: related works}
To recap Section \ref{sec: intro related works}, the two core distinctions that we present from existing work are in:
\begin{itemize}
    \item Formulating a definition of multi-label fairness that is grounded in the utility model that underlies the final serving layer.
    \item Connecting the two key mechanisms for industry-scale recommendation systems - filtration and ranking, and demonstrating how biases in either step can lead to downstream fairness gaps in user utility. 
\end{itemize}

We organize our literature review as they relate to these two points separately. In terms of connecting the notions of industrial ML systems with algorithmic frameworks for fairness and utility, \citet{Ekstrand_2022} represents the closest related work through a comprehensive overview of industry-scale modeling systems. The authors recognize the distinction between short-term proxies and user utility, as well as various fairness definitions in ranking. While they identify fairness issues beyond the ML modeling stage---such as in the candidate retrieval layer and behavioral distribution shifts across groups---they fall back on traditional individual model fairness and do not address the practical compositional nature of the problem.

\subsection{Recap of Fairness in Multi-Label Settings}
Previous works have recognized that industrial systems often comprise multiple models, each predicting different which are proxies of downstream outcomes (e.g., "clicks" as a proxy of engagement, "likes" as a proxy of preference). In these settings, the authors typically develop a multi-label analogs of existing fairness definitions such as equalized odds (\cite{MultiGroupCompositionalFairness}). Specifically, this entails defining a "composite label" $Y_{C}$ as a function of the individual labels $Y_{1}, \ldots, Y_{K}$, each of which represent a different aspect of the goodness of an item (and similarly for the composite prediction $\hat{Y}_{C}$), and then applying a standard single fairness definition on the composite label and prediction. A key motive behind our research was our opinion that the proposed "composite" labels do not adequately reflect how predictions are served. To give some examples, \citet{Dwork2018FairnessUC} and \citet{PraticalCompositionalFairness} suggest having the composite label $Y_{C}$ as the product over the individual labels $Y_{C} = \prod_{k} Y_{k}$ (and similarly for the composite prediction). On the other hand, \citet{MultiGroupCompositionalFairness} suggests using the maxima such that $Y_{C} = \max_{k} Y_{k}$. From a user utility perspective, the interpretation of the product means that the item must be qualified in \textit{all} aspects to be useful for the user, while the interpretation of the maxima means that a singular good aspect makes the item useful for the user. Our definition of the composite label presented in Section \ref{subsection: preference misspecification} as a weighted sum $Y_{C}=\sum_{k} \alpha_{k}^{*}Y_{k}$ therefore takes a natural middle ground, where users have (potentially heterogeneous) preferences over different aspects of an item. As mentioned in Figure \ref{fig:System}, this is indeed the mechanism used across multiple industrial recommendation systems. \citet{diciccio2022predictive} studies a closely related setting by ensuring that the weighted sum of predictions is calibrated with respect to the weighted sum of labels. This is the same setup that we analyze, with the key difference being that \citet{diciccio2022predictive} addresses fairness for the items being ranked, rather than that of the viewer, which is our focus in this paper.

\subsection{Recap of Fairness in Pipelines}
Several fairness works have been motivated by the use of pipeline-style (also termed "sequential") systems in the real world to make decisions. For instance, resume filtering\footnote{We term this situation as "resume filtering", where individuals are being filtered, to disambiguate the situation that we are in, which is when jobs are filtered and shown to a user. In the former situation, the fairness is with respect to the candidates while in the latter, the fairness is with respect to the viewer.} and promotion candidacy are commonly framed as a multi-step filter process, where a decision-maker reduces the candidate pool at each step until they end up with a group of desired size. \citet{bower2017fairpipelines} is an early work in this category and exactly studies fairness in pipeline systems, where a pipeline is fair if the final outcome obeys equal opportunity. In it, the authors illustrate that a pipeline constructed of models that are stand-alone fair may not be fair with respect to the final outcome. \citet{dwork2020individual} analyzes a similar setting, but instead assesses notions of individual fairness rather than group fairness and makes similar conclusions. \citet{blum2022multi} studies the same setting, but focuses on formulating a constrained optimization approach to optimize for both performance and fairness.

In our work, we connect the retrieval/selection pipeline with the serving layer as an extension of the above ideas. This paradigm introduces the perspective that pipeline systems can be a bottleneck for fairness specifically because they can be disconnected from the objectives of the serving layer. Namely, whereas the serving layer may optimize heavily for one of the labels $Y_{k}$, the retrieval pipeline may have been optimized for an entirely different label. In other words, the pipeline step could be largely disconnected from the actual downstream proxies and the users' heterogeneous preferences for those proxies. If there is only one single aspect of relevance (i.e. single-label case), then the prior research of fairness in pipelines directly applies in our setting. However, when items have multiple aspects of qualification (multi-label case), the quality of a retrieval mechanism becomes more nuanced, as we describe in Section \ref{subsection: Retrival distribution shift}. 

\section{Appendix / Proofs}
\subsection{Proof of Lemma \ref{lemma: optimal alpha}}
\begin{proof}
First we observe that by definition, the optimal utility is realized when we pick the best retrieved item with respect to the true expected outcomes conditional on the predictions $\mathbb{E}[Y_{k}|f_{1}(X,Z^{j}),\ldots,f_{K}(X,Z^{j})]$ and true user preferences $\alpha^{*}_{k}$. Hence, the best item $j^{*}$ is the solution to the following, where without loss of generality we only need to consider the selected items where $I(X,Z^{j})=1$, which we denote as $1,\ldots, m^{+}$:
\[
    j^{*} := \mbox{argmax}_{1 \le j \le m^{+}}
    \mathbb{E}\left[ 
    \sum_{k=1}^K \alpha_{k}^{*} Y_{k}^{j} \big| f_{1}(X,Z^{j}),\ldots,f_{K}(X,Z^{j})
    \right]
\]
Now we show that this item $j^{*}$ is indeed selected whenever $\alpha= c\cdot \alpha^{*}$ and $c > 0$ by plugging these values into the selection function Eq. \ref{eq: argmax selection} (and denoting $\hat{j}(\alpha)$ as the retrieved item when using SPR coefficients $\alpha$).

\begin{align*}
    \hat{j}(\alpha^{*}) & := \mbox{argmax}_{1 \le j \le m^{+}} \sum_{k=1}^K \alpha_k f_k(X, Z^j) \\
    & =\mbox{argmax}_{1 \le j \le m^{+}} \sum_{k=1}^K c\cdot \alpha_{k}^{*} \cdot \mathbb{E}\left[Y_{k}^{j} \mid f_{1}(X, Z^{j}), \ldots, f_{K}(X, Z^{j})\right]  \\
    & = \mbox{argmax}_{1 \le j \le m^{+}}
    \mathbb{E}\left[ 
    c \sum_{k=1}^K \alpha_{k}^{*} Y_{k}^{j}  \big| f_{1}(X,Z^{j}),\ldots,f_{K}(X,Z^{j})
    \right] 
\end{align*}

Where in the second line, we utilized the assumption that the model is calibrated. In the last line, we rely on the fact that if $c > 0$, the argmax of the last line is still the same as $j^{*}$. 

Next, to show that if we select $\alpha' \neq c\cdot \alpha^{*}$ then $U(X,I,f,\alpha',\alpha^{*}) \leq U(X,I,f,c\cdot\alpha^{*},\alpha^{*})$, we use the assumption that any item can be selected a.s. and that $f_{k}(X,Z^{j})$ can take on any value in its range to demonstrate that we can always construct cases where using $\alpha'$ will lead to selection of a suboptimal item. To first provide some intuition for why $\alpha'$ can be suboptimal, consider changing a single coordinate such that $\alpha_{k}'=c\cdot \alpha_{k}^{*}+\epsilon_{1}$. Then the selected item will be based on:
\[
\hat{j}(\alpha') := \mbox{argmax}_{1 \le j \le m^{+}}
    \mathbb{E}\left[ 
    \epsilon Y_{k} + c \sum_{k=1}^K \alpha_{k}^{*} Y_{k}^{j}  \big| f_{1}(X,Z^{j}),\ldots,f_{K}(X,Z^{j})
    \right]
\]
From this, we can see that the argmax now considers an extra term $\epsilon Y_{k}$ which is disconnected the true preferences of label $k$. The key intuition is therefore that if an item has a label with very low true preference has high expectation, then it may be selected over an item of overall higher quality. We proceed formally via proof by contradiction.

By our assumption, we have a nonzero probability of encountering the following list with two items $Z^{-}$ and $Z^{+}$. We construct a list so that both items have $\mathbb{E}[Y_{k}|f]=b_{k}$ (where we use the shorthand $f$ in the conditioning to denote the set of predictions for an item) except at indices $i$ and $j$, where for scalars $e_{j}^{+}, e_{i}^{-}, e_{j}^{-} \in \mathbb{R}$, we have that:
\[ E[Y^{+}|f] = \{b_{1}, b_{2}, \ldots, b_{i}, b_{j}+e_{j}^{+}, \ldots, b_{K} \} \]
\[ E[Y^{-}|f] = \{b_{1}, b_{2}, \ldots, b_{i}+e_{i}^{-}, b_{j}+e_{j}^{-}, \ldots, b_{K} \} \]
We continue the construction by letting $Z^{+}$ have higher ground truth relevance. That is:

\begin{align}
    & \mathbb{E} \left[ \sum_{k}^{K} \alpha_{k}^{*} Y^{+}_{k} \right] > 
    \mathbb{E} \left[ \sum_{k}^{K} \alpha_{k}^{*} Y^{-}_{k} \right] \nonumber \\ 
    & \implies \sum_{k}^{K} \alpha_{k}^{*} b_{k} + \alpha_{j}^{*}e_{j}^{+} > 
    \sum_{k}^{K} \alpha_{k}^{*} b_{k} + \alpha_{i}^{*}e_{i}^{-} + \alpha_{j}^{*}e_{j}^{-} \nonumber\\ 
    & \iff \alpha_{j}^{*}e_{j}^{+} > \alpha_{i}^{*}e_{i}^{-} + \alpha_{j}^{*}e_{j}^{-} \nonumber\\ 
    & \iff (e_{j}^{+} - e_{j}^{-}) > \frac{\alpha^{*}_{i}}{\alpha^{*}_{j}} e_{i}^{-} \label{cond: e}
\end{align}

In the last line, division is permitted as $\alpha^{*}$ are positive by construction. Now suppose that we serve recommendations using preferences that diverge from $c\cdot \alpha^{*}$ by using $\alpha' = c\cdot \alpha^{*} + \beta d$ where we have $\beta d \neq c\cdot \alpha^{*}$ (i.e., $\alpha'$ is not in the same direction as $\alpha^{*}$) and $\beta d > -c \cdot \alpha^{*}$ (since preferences cannot be negative). We assume that $\alpha'$ is an optimal solution, which implies that there does not exist a set of $e_{j}^{+}, e_{j}^{-}, e_{i}^{-}$ both satisfying inequality \ref{cond: e} and leads to $Z^{-}$ getting picked (therefore yielding suboptimal utility). Having $Z^{+}$ be selected requires that

\begin{align*}
    & \mathbb{E} \left[ \sum_{k}^{K} \alpha_{k}' Y^{+}_{k} \right] > 
    \mathbb{E} \left[ \sum_{k}^{K} \alpha_{k}' Y^{-}_{k} \right] \\ 
    & \implies \sum_{k}^{K} (c\cdot \alpha_{k}^{*} + \beta d_{k}) b_{k} + (c\cdot \alpha_{j}^{*} + \beta d_{j}) e_{j}^{+} \\ 
    & \qquad > \sum_{k}^{K} (c\cdot \alpha_{k}^{*} + \beta d_{k}) b_{k} + (c\cdot \alpha_{i}^{*} + \beta d_{i}) e_{i}^{-} + (c\cdot \alpha_{j}^{*} + \beta d_{j}) e_{j}^{-} \\ 
    & \iff (c\cdot \alpha_{j}^{*} + \beta d_{j}) e_{j}^{+} >  (c\cdot \alpha_{i}^{*} + \beta d_{i}) e_{i}^{-} + (c\cdot \alpha_{j}^{*} + \beta d_{j}) e_{j}^{-} \\
    & \iff e_{i}^{-} < (e_{j}^{+} - e_{j}^{-}) 
        \left( \frac{c\cdot \alpha_{j}^{*} + \beta d_{j}}{c\cdot \alpha_{i}^{*} + \beta d_{i}}  \right)
\end{align*}

Where in the first line, we cancel out the large sum and leave the extra terms involving $e_{j}^{+}, e_{j}^{-}, e_{i}^{-}$ and in the second and third lines we can safely divide by $c\cdot \alpha_{j}^{*} + \beta d_{j}$ since by construction it is positive. Now we apply inequality $\ref{cond: e}$ on the last line and let $(e_{j}^{+} - e_{j}^{-}) = \frac{\alpha^{*}_{i}}{\alpha^{*}_{j}} e_{i}^{-} + \epsilon$ for some $\epsilon > 0$. This gives us the following inequality:

\begin{align*}
    & e_{i}^{-} < \left(\frac{\alpha^{*}_{i}}{\alpha^{*}_{j}} e_{i}^{-} + \epsilon \right)
        \left( \frac{c\cdot \alpha_{j}^{*} + \beta d_{j}}{c\cdot \alpha_{i}^{*} + \beta d_{i}}  \right) \\
    & \iff 0 < e_{i}^{-} \left[ \frac{\alpha^{*}_{i}}{\alpha^{*}_{j}} \cdot \frac{c\cdot \alpha_{j}^{*} + \beta d_{j}}{c\cdot \alpha_{i}^{*} + \beta d_{i}} -1  \right] + \epsilon \left( \frac{c\cdot \alpha_{j}^{*} + \beta d_{j}}{c\cdot \alpha_{i}^{*} + \beta d_{i}}  \right)
\end{align*}

At this point, observe that since $\epsilon > 0$ and $c\cdot \alpha + \beta d > 0$ for both $d \in \{d_{i}, d_{j}\}$ and hence the second term on the right-hand side of the inequality is positive. However, if the term multiplied with $e_{i}^{-}$ is nonzero, we can use our construction to drive $e_{i}^{-}$ upwards (by widening the upper bound gap of $e_{j}^{+} - e_{j}^{-}$) or downwards to make the inequality false. Making the term next to $e_{i}^{-}$ zero requires that $d_{i}=\beta \alpha^{*}_{i}$ and $d_{j}=\beta \alpha^{*}_{j}$, but this is false by assumption (since we assumed that the $d$ vector is in a different direction of the optional $\alpha^{*}$ vector) and we have thus reached a contradiction for an arbitrary choice of $\beta$, $d_{i}, d_{j}$ (including if one of them equals zero). Putting these statements together, we have shown that $\alpha^{*}$ is a maximizer of the utility function.
\end{proof}

\subsection{Proof of Theorem \ref{lemma: alpha shift}}
We show that the expected difference in groupwise utility over the shared distribution (under fair, calibrated models) can be upper bounded as follow:
\begin{proof}
    \begin{align*}
    \mathbb{E}_{S_{X}} & \left[ U_{1}(X, I, f, \alpha, \alpha^{*}) - U_{0}(X, I, f, \alpha, \alpha^{*}) \right] \\
    &= \mathbb{E}_{S_{X}} \left[ \mathbb{E}\biggl[ \sum_{k} \alpha_{k}^{*}(1) Y_{k}^{\hat{j}} - \alpha_{k}^{*}(0)Y_{k}^{\hat{j}} \big| X \biggr] \right] \\
    &= \mathbb{E}_{S_{X}} \biggl[ \mathbb{E} \biggl[ \sum_{k} \alpha^{*}_{k}(1)(Y_{k}^{\hat{j}}-f_{k}(X,Z^{\hat{j}})) - \alpha_{k}^{*}(0)(Y_{k}^{\hat{j}}-f_{k}(X,Z^{\hat{j}})) \\
    & \qquad + (\alpha_{k}-\alpha_{k}^{*}(0)) f_{k}(X,Z^{\hat{j}}) +(\alpha_{k}^{*}(1)-\alpha_{k}) f_{k}(X,Z^{\hat{j}}) \big| X \biggr] \biggr] \\
    & \leq \mathbb{E}_{S_{X}} \biggl[ \mathbb{E} \biggl[ \sum_{k} (\alpha^{*}_{k}(1)-\alpha_{k}^{*}(0) (Y_{k}^{\hat{j}}-f_{k}(X,Z^{\hat{j}})) \\
    & \qquad + |\alpha_{k}-\alpha_{k}^{*}(0)| f_{k}(X,Z^{\hat{j}}) + |\alpha_{k}-\alpha_{k}^{*}(1)| f_{k}(X,Z^{\hat{j}}) \big| X \biggr] \biggr] \\
\end{align*}
\end{proof}
In the third line, we add and subtract terms and then collect them. In the last line, we utilize that $x \leq |x|$. 

\subsection{Proof of Theorem \ref{lemma: CR dist shift}}
\begin{proof}
    By definition, we have:
    \[
    Q_{m}(I(X,Z), Y) = \mathbb{E}\left[\max_{j \in \{1, \ldots, m\}} \left( I(X, Z_j) \cdot \sum_{k=1}^{K} Y_{k}^{j} \right)\right].
    \]
    Since we assumed that the serving model is optimal as we are using calibrated models and know the true preferences w.r.t. each label, this implies that the expected utility optimized with respect to the best of the selected items. Since $Q_{m}(I(X,Z), Y)$ represents the expected maximum $\gamma$-good item retrieved, without loss of generality we can say that the expected utility on the distribution $X|G=1$ is upper bounded by $\gamma > 0$ such that:
    \[
    0 \leq \mathbb{E}_{1} \left[ U_{1}(X, I, f, \alpha, \alpha^{*}) \right] \leq \gamma \cdot \lvert \alpha^{*}_{k}(0) \rvert_{\infty}
    \]
    By similar reasoning, the expected utility on the distribution $S_{X}$ is upper bounded as:
    \[
    0 \leq \mathbb{E}_{S_{X}} \left[ U_{1}(X, I, f, \alpha, \alpha^{*}) \right] \leq (\gamma-\epsilon) \cdot \lvert \alpha^{*}_{k}(0) \rvert_{\infty}
    \]
    Putting these bounds together to analyze the difference in the best case expected utility across the distributions, we get that:

    \begin{align*}
    \sup_{f,\alpha} \mathbb{E}_{1} \left[ U_{1}(X, I, f, \alpha, \alpha^{*}) \right] & - \sup_{f,\alpha}  \mathbb{E}_{S_{X}} \left[ U_{1}(X, I, f, \alpha, \alpha^{*}) \right] \\
    & = \gamma \cdot \lvert \alpha^{*}_{k}(1) \rvert_{\infty} - (\gamma-\epsilon) \cdot \lvert \alpha^{*}_{k}(1) \rvert_{\infty} \\
    & = \epsilon \cdot \lvert \alpha^{*}_{k}(1) \rvert_{\infty} \\
    & \geq \epsilon \cdot \min_{k} \alpha^{*}_{k}(1)
    \end{align*}
    
    The last inequality follows from the fact that $\lvert \alpha^{*}_{k}(1) \rvert_{\infty} \geq \min_{k} \alpha^{*}_{k}(1)$. Intuitively, this means that even if we optimize the model and serving coefficients, because there is a discrepancy in the quality of the candidate retrieval on at least one of the labels $Y_k$ across distributions $X|G=1$ and $S_x$, this will propagate through the pipeline in the form of a utility gap across groups.

\end{proof}

\section{Appendix / Data and Experiment Details} \label{appendix: data}

In this section, we first illustrate the diversity of the datasets by showing the base performance metrics of each model as well as the utility and DER surfaces. Then, we provide details on how each dataset and model was constructed. 

\subsection{Model performance}
As each dataset has two labels, we show the model performance on each label for the overall and for each group. We keep each dataset at a split of 80/20 across the two groups. This is to simulate the primary fairness concern of our work which is that optimizing for global utility in the presence of heterogeneous preferences and imbalanced representation will lead to disproportional benefits for one group.

\begin{table}[h!]

    \parbox{.45 \linewidth}{
        \centering
    {\begin{tabular}{llll}
        \toprule
        Group & Prevalence & $Y_{1}$ AUC & $Y_{2}$ AUC \\
        \midrule
        All   & 100\%      & 0.995    & 0.996    \\
        0     & 80\%       & 0.996    & 0.996    \\
        1     & 20\%       & 0.995    & 0.996    \\
        \bottomrule
   \end{tabular}}
        \caption{Synthetic Model Performance \label{tbl: synth performance}}
    }
    \hfill
    \parbox{.55 \linewidth}{
        \centering
    {\begin{tabular}{llll}
        \toprule
        Group & Prevalence & $Y_{1}$ AUC & $Y_{2}$ AUC \\
        \midrule
        All   & 100\%      & 0.782    & 0.929    \\
        0     & 80\%       & 0.780    & 0.927    \\
        1     & 20\%       & 0.790    & 0.934    \\
        \bottomrule
   \end{tabular}}
        \caption{MovieLens Model Performance \label{tbl: movelens performance}}}

\end{table}

\begin{table}[h!]

    \parbox{.45 \linewidth}{
        \centering
    {\begin{tabular}{llll}
        \toprule
        Group & Prevalence & $Y_{1}$ AUC & $Y_{2}$ AUC \\
        \midrule
        All   & 100\%      & 0.827    & 0.869    \\
        0     & 80\%       & 0.830    & 0.854    \\
        1     & 20\%       & 0.826    & 0.872    \\
        \bottomrule
   \end{tabular}}
        \caption{Modcloth Model Performance \label{tbl: modcloth performance}}
    }
    \hfill
    \parbox{.55 \linewidth}{
        \centering
    {\begin{tabular}{llll}
        \toprule
        Group & Prevalence & $Y_{1}$ AUC & $Y_{2}$ AUC \\
        \midrule
        All   & 100\%      & 0.861    & 0.830    \\
        0     & 80\%       & 0.857    & 0.829    \\
        1     & 20\%       & 0.865    & 0.830    \\
        \bottomrule
   \end{tabular}}
        \caption{Electronics Model Performance \label{tbl: electronics performance}}}

\end{table}

\subsection{Utility and DER surface}

\begin{figure}[h!]
    \centering
    \includegraphics[width=\textwidth]{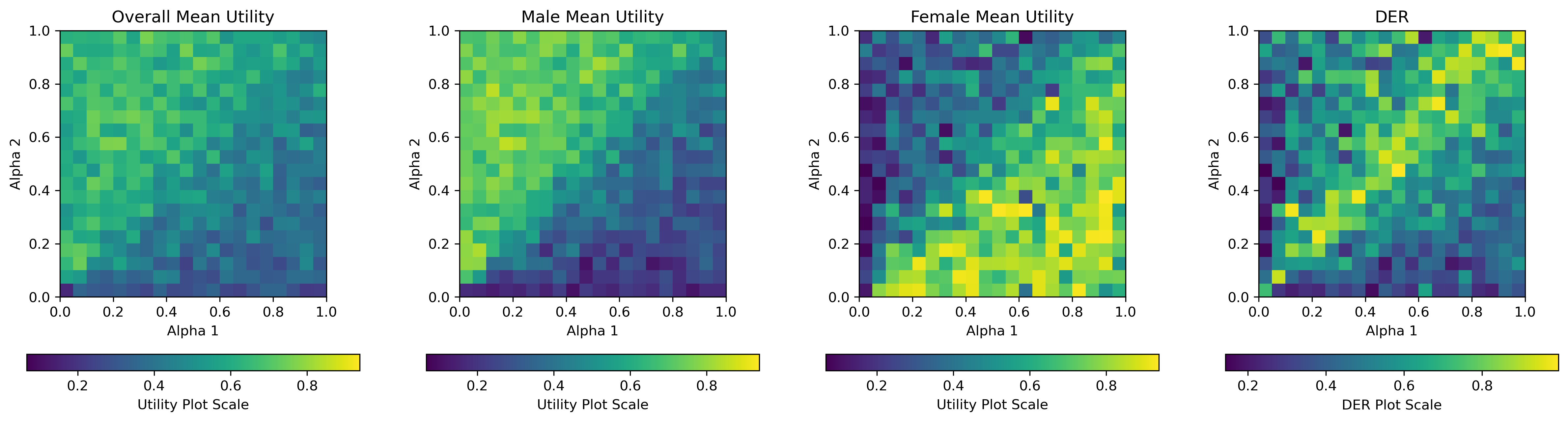}
    \caption{Utility and DER surface - Synthetic data}
    \label{fig:UD surface - synth}
\end{figure}

\begin{figure}[h!]
    \centering
    \includegraphics[width=\textwidth]{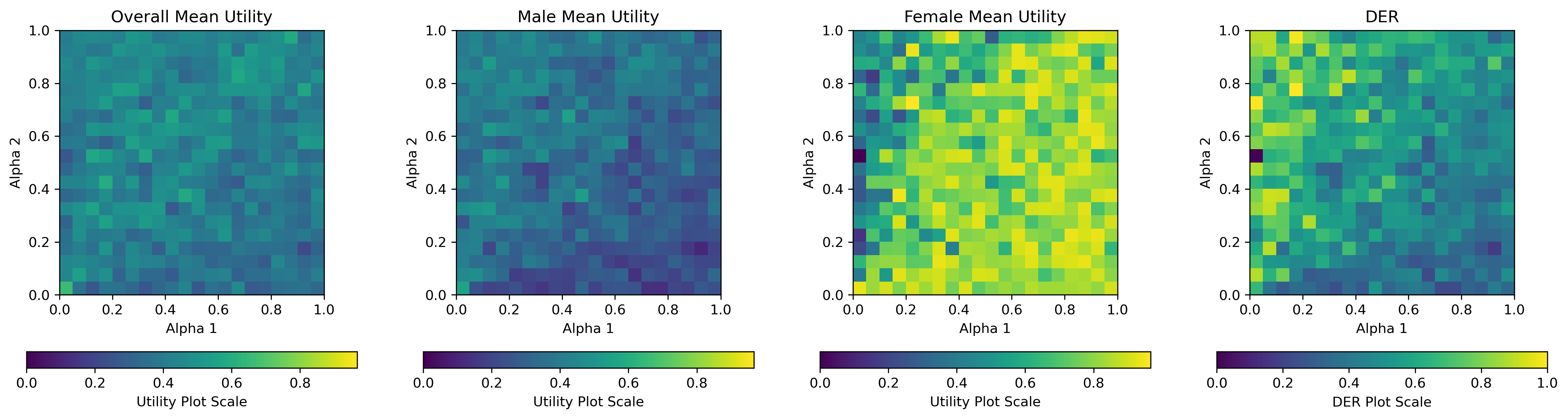}
    \caption{Utility and DER surface - Movielens data}
    \label{fig:UD surface - Movielens}
\end{figure}

\begin{figure}[h!]
    \centering
    \includegraphics[width=\textwidth]{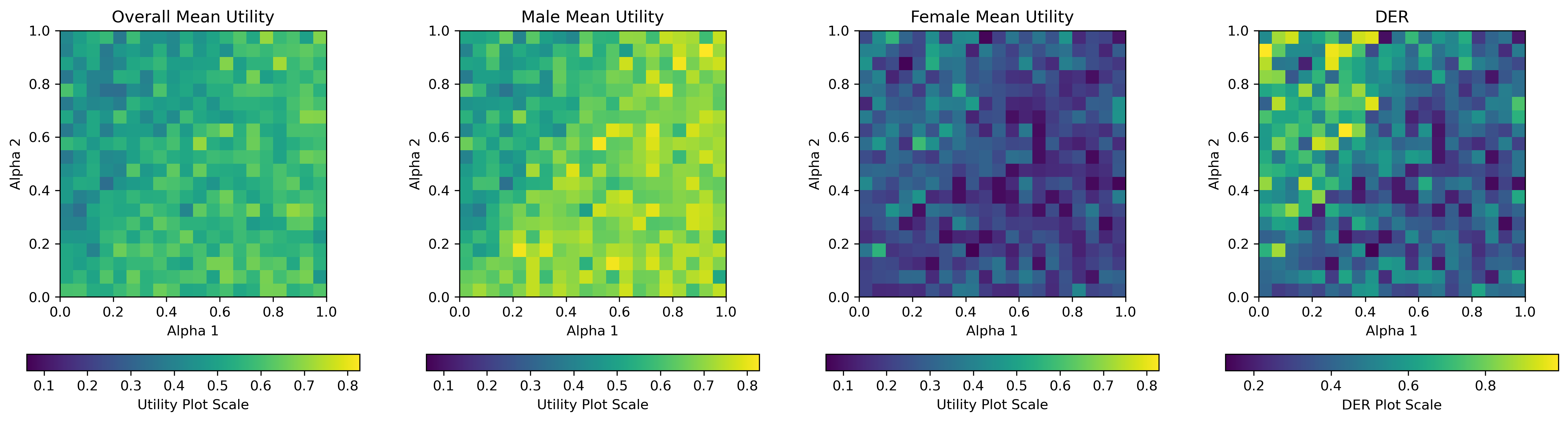}
    \caption{Utility and DER surface - Modcloth data}
    \label{fig:UD surface - Modcloth}
\end{figure}

 Loosely speaking, the hardness of this multi-objective Bayesian Optimization problem comes down to how hard it is the find the globally optimal utility and DER separately, and how close or far those points are from each other (which makes finding the tradeoff difficult). These factors in turn depend on several factors, including the sparsity of positive labels, the noisiness of the labels, and model's predictive ability. For example, Figure \ref{fig:UD surface - synth} shows that the optimal multi-optjective solution in the clean Synthetic data case is likely any point on the diagonal where the two $\alpha_{k}$ values are equal. However, the optimal utility in the Movielens data is a much noisier surface as shown in Figure \ref{fig:UD surface - Movielens}. The Modcloth data in Figure \ref{fig:UD surface - Modcloth} showcases an instance where the region with the optimal DER is in the opposite corner as the optimal utility. Lastly, the Electronics data is characterized by a relatively flat utility and DER surface, where any random point could feasibly yield relatively high utility and DER. This explains why both EI and Fair EHVI did not beat random search by a wide margin. In fact, EI had an average performance below random search in our experiments.

\begin{figure}[h!]
    \centering
    \includegraphics[width=\textwidth]{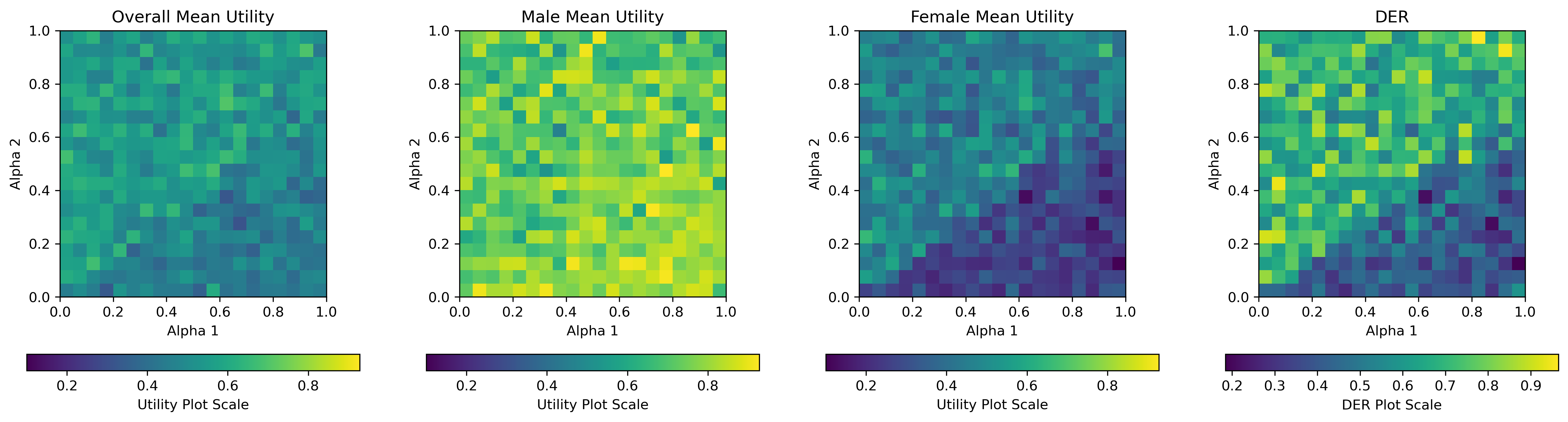}
    \caption{Utility and DER surface - Electronics data}
    \label{fig:UD surface - Electronics}
\end{figure}

\subsection{Data Generation}

We now explain how we processed each dataset, created models, and processed the scores to create labels. We start with the MovieLens dataset, being the most complex model as it required hand-crafting features. Then we describe the Modcloth and Electronics datasets, which we processed by modifying the source code from \citet{wan2019addressingmarketingbiasproduct}. Finally, we end with the Synthetic dataset as it is identical in the mechanics to the above processes, but significantly simpler to implement. In particular, the general process of processing these datasets is sampling synthetic negatives, constructing a model, and placing the data into a batch serving and scoring mechanism. The last step (common to all datasets) is designed to simulate a real-world recommendation system. In it, we simulate a batch of users coming onto the platform, where we fetch 20 items for each user, rank them based on predicted preferences, and finally obtaining feedback in terms of user utility on the best item (as stated in our framework).

In terms of the models used, we use random forests for Synthetic and MovieLens datasets and neural networks for the Modcloth and Electronics datasets. We emphasize that the specific model used does not matter beyond the performance of the model, which we disclosed above. This is because our algorithmic innovation is strictly on the adaptive experiment side using BO, which effectively treats the model as a black-box system and only sees the scores and the labels.

\subsubsection{MovieLens}
We use the MovieLens dataset. The original task in MovieLens is to predict movie ratings given user ID, item ID, and one-hot encoded genre data about each movie that a user watched. We transform this into a multi-task, multi-stage, preference learning problem. We prepare the data generation process as follows:
\begin{enumerate}
    \item Generate a new label $y_{watch}$, which represents if a user watched a movie. We generate random synthetic negatives in the data to create this label. We let the other label be $y_{liked}$ based on if the user watched the movie and gave it a rating of 3 or higher.
    \begin{itemize}
        \item Trivial solutions can be optimal if the labels have a certain dependency structure (e.g. having perfect correlation), for example if $y_{like}=1$ is possible only if $y_{watch}=1$. Hence even for synthetically generated negative movies that are not watched, we impute a label by first learning the patterns of likes for the movies that are watched and then applying that like prediction (Scikit-learn (\cite{scikit-learn}) random forest) model to the movies that are not watched.
        \item When doing this, we also scramble the gender assignment in the data as we assume no $Y|X$ distribution shift across groups. This is to remove $Y|X$ distribution shifts from the data which are outside the scope of our framework. 
    \end{itemize}
    \item Generate user features $X$ for both tasks by taking the historical rolling average of the genres of movies that a user watched. E.g. if you watched 2 documentaries and 1 action movie in a rolling window of 3, your feature matrix for those genres would look like $[2/3, 1/3]$. These are used alongside the genre information of the movie being scored which represent the item features $Z^{j}$. 
    \item Split the data. We take a random 20\% of the data for the models and use the other 80\% for the BO process. For the models, we train one model for each label that maps $f_{k}: X \rightarrow [0,1]$ to model $y_{k}$ with probabilistic predictions. We train the scoring model via a Scikit-learn random forest classifier with default parameters and $n=300$ trees. A random forest was chosen as in the authors' experience, these models work well off-the-shelf with minimal tuning.
        \begin{itemize}
            \item We tested other off-the-shelf scikit-learn models but found that this provided the best performance in terms of ROC AUC. We did not focus on any hyperparameter tuning as default settings provided good results.
        \end{itemize}
    \item Create batch-sampling, batch-scoring, and batch-utility mechanisms: The point is to simulate random users coming onto the site, getting some recommendations, we score them with $f_{k}$, serve recommendations using $\alpha$, and then observe rewards.
        \begin{itemize}
            \item The user's $u$ true preference of each item $i$ shown is $\sum_{k} \alpha^{*}_{k}y_{k}$. Importantly, \textit{$\alpha^{*}$ depends on the user's demographics.}
            \item The predicted preference is $\sum_{k} \alpha_{k}f_{k}$ where $\alpha$ is the PM's guess of user preference
            \item Compute the true preference of the highest scoring predicted item, that is the user's utility
            \item Repeat the above steps for all users.
        \end{itemize}
\end{enumerate}

\subsubsection{Modcloth and Electronics}
The Modcloth and Electronics datasets are both e-commerce datasets. They provide data about user demographics (which the authors thoughtfully imputed) as well as user ratings of items that they purchased (clothing and electronics, respectively). We largely rely on the source code provided in \citet{wan2019addressingmarketingbiasproduct} for our purposes with some modifications which we detail below. In it, the authors model the different characteristics such as "fit" and "rating" using a collaborative filtering neural network model. Specifically, the model involves storing embeddings for each user and item, then training the embeddings by using the dot product to make predictions. Our modifications are as follows:
\begin{enumerate}
    \item For both datasets, we need synthetic negatives to represent samples presented to the user that they would not have purchased. The original authors provide a method of generating these samples by creating a distribution over user and item IDs such that the probability of sampling a pair (as a negative) is 0 if the user/item pair exists in the data with a rating of 4 or higher. We use this exact mechanism. Then to impute the labels, we sample from a probability distribution where the density of each rating is shifted down by 1 (e.g., the probability of a 4 for a negative item is the probability of a 3 for a non-negative item). This gives us "viewed" binary labels.
    \item We translate the other labels "rating" and "fit" into binary labels as well based on thresholds. We then use these labels to train a collaborative filtering model. The main difference between our model and the original is that turn it into a multitask problems where each task shares user/item embeddings and also has a task-specific layer. Additionally, we change the loss from mean squared error to binary cross-entropy. 
    \item Lastly, for both datasets, we observed that the labels were positively correlated to begin with (potentially due to the built-in negative sampling mechanism). We found this to be unrealistic in that generally, an ML system would focus on predicting signals that capture different characteristics of an item. In other words, an industrial application is unlikely to increase the serving complexity by building multiple models that essentially capture the same signal. Hence, we flipped the correlation of the labels and predictions by taking the \textit{opposite} of one of the labels ("viewed" for Modcloth and "rating" for Electronics). The intuition is that now, the labels capture different aspects of "goodness" of the item such that being "good" in one label does not necessarily mean it is "good" in another label. 
\end{enumerate}

Beyond the steps above, the batch-sampling, batch-scoring, and batch-utility mechanisms are identical to the functions we constructed in the MovieLens dataset.

\subsubsection{Synthetic Data}
For the synthetic data, we utilize the Scikit-learn Python package to generate two classification datasets independently. We randomly assign groups to the data and then again train the scoring model via a random forest with default parameters and $n=300$ trees. As there are no user/session IDs in the synthetic data, we simply collect a batch of random samples at each iteration. The batch sampling mechanism and utility computation that we use in this dataset is identical to the last step to that of the MovieLens data.

\subsection{EI and Fair EHVI Details}
We use the Botorch implementations of parallel EI (\texttt{qExpectedImprovement}) and EHVI (\texttt{qExpectedHypervolumeImprovement} for the solution implementation and comparisons. Specifically, the implementation provided in \citet{balandat2020botorch} and \citet{daulton2020ehvi} allow for parallel EHVI (called q-EHVI). In essence, this algorithm addresses the nonconvexity of the problem by sampling, optimizing, and yielding $q$ points that are expected to yield higher hypervolume. We use $q=10$ and considered three strategies for picking a single point to use.
\begin{enumerate}
    \item Pick the point that has the highest expected DER improvement
    \item Pick the point that has the highest utility improvement
    \item Pick a random point (since all points are expected to improve the hypervolume of the Pareto frontier)
\end{enumerate}

We found that the first strategy worked the best. For random search, we simply generate random alphas to use in each iteration.

\section{Appendix / Role of Unobservable Outcomes} \label{appendix: unobservables}
As mentioned in Section \ref{sec: Framework}, one thing we did not explicitly consider in this framework is the role of unobservable outcomes. In practical settings, unobservable outcomes may be factors that are impossible to capture, but nonetheless affect the user's utility (e.g., the user's "agreement" with the company's mission in a job recommendation, the user's opinion of the posting description). They could also refer to things that are technically capturable, but left out for simplicity in the serving framework (e.g. transformations or products of click propensity, view propensity). From an interpretability standpoint, these omissions are reasonable. While one can easily place relative values on things (e.g. "a click means twice the utility to a user compared to a view"), doing so becomes harder when those factors become more abstract. Hence, unobservable outcomes are inevitably in the picture even if they are not directly modeled.

In the context of our framework, one cause of utility gap could be that of the excluded unobservable outcomes, one group places much higher weight on them than another group. Hence excluding them hurts one group more than the other. In the industrial setting, mitigating this requires product-based intervention in similar vein as discussed in \citet{caiAdaptive}. That is, we would recommend the business to conduct user research and segment the analysis across groups to better understand what preferences the users actually have versus what is currently modeled and served. From this lens, our framework can be utilized to quantitatively understand the benefits of product research. 
\newpage
\bibliographystyle{abbrvnat}
\bibliography{references}

\end{document}